\documentclass[final]{cvpr}

\usepackage{times}
\usepackage{epsfig} 
\usepackage{graphicx} 
\usepackage{amsmath}
\usepackage{amssymb}

\usepackage{float}  
\usepackage{booktabs}
\usepackage{caption}
\usepackage{multirow} 
\usepackage{xcolor}
\usepackage{url}
\usepackage{cite}
\usepackage{subcaption}
\usepackage{placeins}
\usepackage[group-separator={,}]{siunitx}
\usepackage[utf8]{inputenc}

\usepackage[pagebackref=true,breaklinks=true,colorlinks,bookmarks=false]{hyperref}

\newcommand{\bc}{\mathbf{c}}
\newcommand{\bd}{\mathbf{d}}

\newcommand{\bI}{\mathbf{I}}

\newcommand{\bR}{\mathbf{R}}

\newcommand{\bt}{\mathbf{t}}\newcommand{\bT}{\mathbf{T}}

\newcommand{\bx}{\mathbf{x}}

\newcommand{\bz}{\mathbf{z}}

\newcommand{\bxi}{\boldsymbol{\xi}}

\newcommand{\nE}{\mathbb{E}}

\newcommand{\cD}{\mathcal{D}}

\newcommand{\cV}{\mathcal{V}}

\makeatletter
\DeclareRobustCommand\onedot{\futurelet\@let@token\@onedot}
\def\@onedot{\ifx\@let@token.\else.\null\fi\xspace}
\def\eg{e.g\onedot}

\def\wrt{wrt\onedot}

\def\etal{et~al\onedot}

\makeatother

\newcommand{\boldparagraph}[1]{\vspace{0.2cm}\noindent{\bf #1:}}

\definecolor{darkgreen}{rgb}{0,0.7,0}

\graphicspath{./gfx/}
\renewcommand{\boldparagraph}[1]{\vspace{.1cm}\noindent\textbf{#1:}}
\newcommand{\boldparagraphnovspace}[1]{\vspace{-0.00cm}\noindent\textbf{#1:}}

\newcommand{\zshape}{\mathbf{z}_s}
\newcommand{\zapp}{\mathbf{z}_a}
\newcommand{\ti}{\mathbf{T}_i}
\newcommand{\rv}{\pi_\text{vol}}
\newcommand{\rn}{\pi_{\theta}^\text{neural}}
\newcommand{\comp}{C}
\newcommand{\igen}{\hat{\mathbf{I}}}

\newcommand{\vspaceunderfig}{\vspace{-0.2cm}}
\newcommand{\vspaceundertab}{\vspace{-.2cm}}
\newcommand{\vspaceunderfigext}{\vspace{-0.7cm}}

\def\cvprPaperID{3367} %
\def\confYear{CVPR 2021}

\begin{document}

\title{GIRAFFE: Representing Scenes as\\Compositional Generative Neural Feature Fields}
\author{Michael Niemeyer$^{1,2}$ \quad Andreas Geiger$^{1,2}$\\
$^1$Max Planck Institute for Intelligent Systems, Tübingen \qquad $^2$University of Tübingen\\
{\tt\small \{firstname.lastname\}@tue.mpg.de}
}

\maketitle 

\thispagestyle{empty} %
\pagestyle{empty} %

\begin{abstract}
Deep generative models allow for photorealistic image synthesis at high resolutions. But for many applications, this is not enough: content creation also needs to be controllable. While several recent works investigate how to disentangle underlying factors of variation in the data, most of them operate in 2D and hence ignore that our world is three-dimensional. Further, only few works consider the compositional nature of scenes. Our key hypothesis is that incorporating a compositional 3D scene representation into the generative model leads to more controllable image synthesis. Representing scenes as compositional generative neural feature fields allows us to disentangle one or multiple objects from the background as well as individual objects' shapes and appearances while learning from unstructured and unposed image collections without any additional supervision. Combining this scene representation with a neural rendering pipeline yields a fast and realistic image synthesis model. As evidenced by our experiments, our model is able to disentangle individual objects and allows for translating and rotating them in the scene as well as changing the camera pose.
\end{abstract}

\section{Introduction}
\begin{figure}
    \centering 
        \centering\includegraphics[width=1.\linewidth]{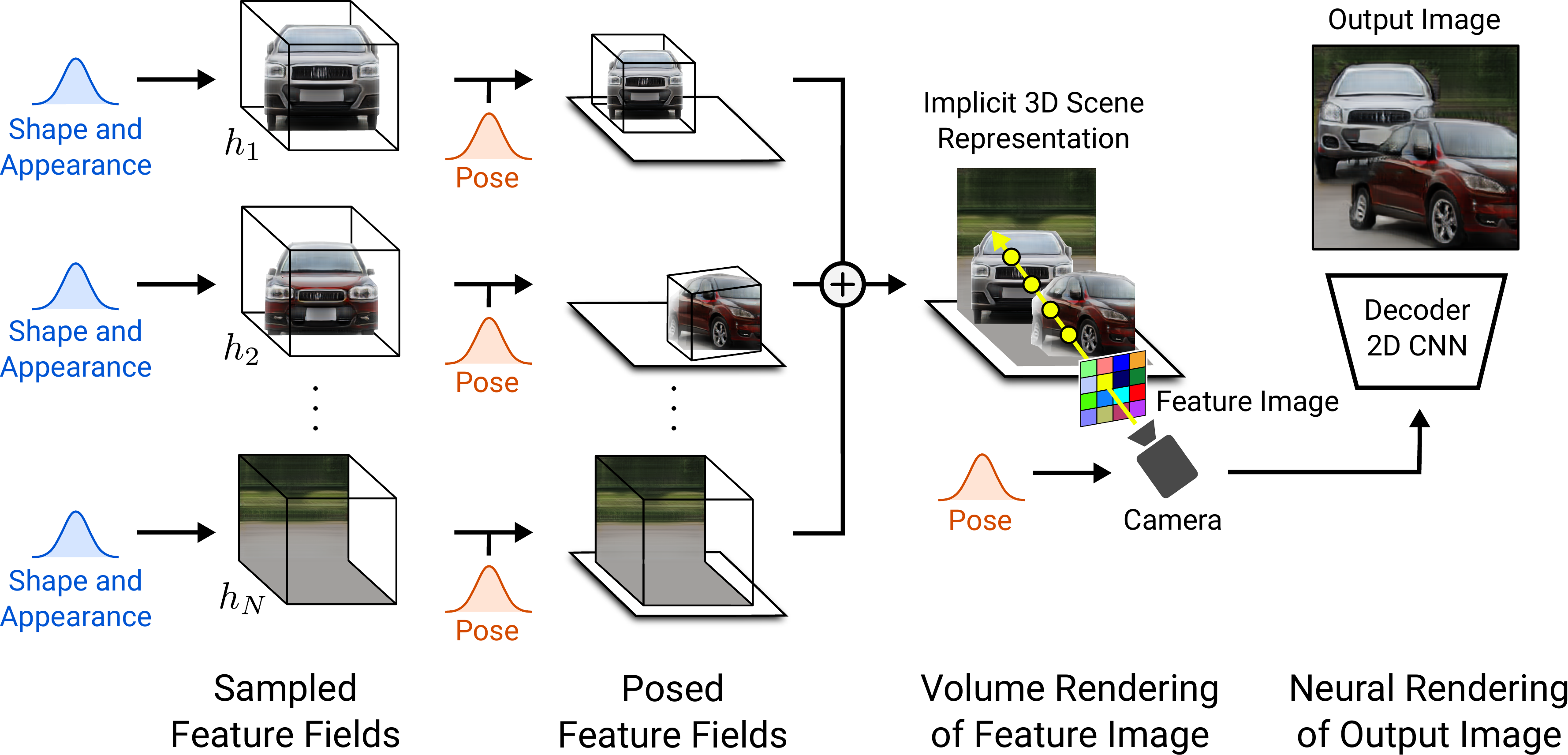}
        \vspace{-.6cm}
        \label{subfig:teaser-a}
    \caption{   
        \textbf{Overview.}
        We represent scenes as compositional generative neural feature fields.
        For a randomly sampled camera, we volume render a feature image of the scene based on individual feature fields.
        A 2D neural rendering network converts the feature image into an RGB image.
        While training only on raw image collections, at test time we are able to control the image formation process \wrt camera pose, object poses, as well as the objects' shapes and appearances.
        Further, our model generalizes beyond the training data, \eg we can synthesize scenes with more objects than were present in the training images.
        Note that for clarity we visualize volumes in color instead of features.
        }
    \label{fig:teaser}
     \vspace{-0.3cm}
\end{figure}
\begin{figure}
  \centering 
    \begin{subfigure}[t]{1.\linewidth}
      \centering\includegraphics[width=\linewidth]{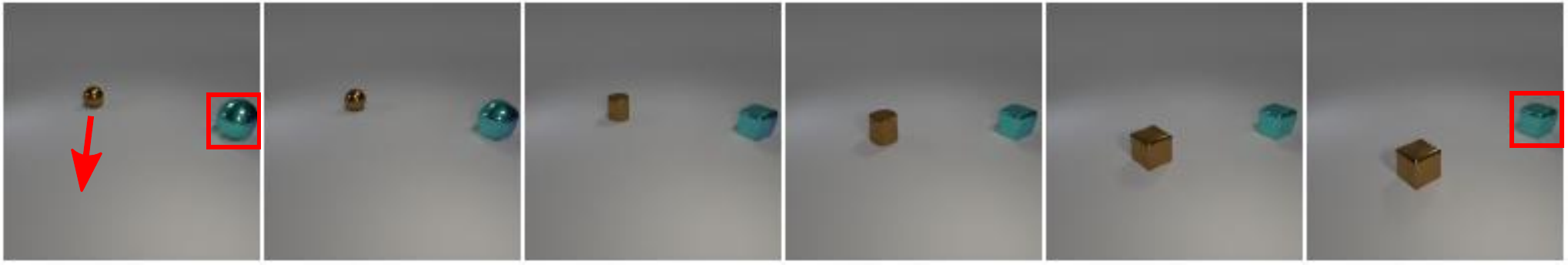}
      \caption{Translation of Left Object (2D-based Method~\cite{Peebles2020ECCV})}
      \label{fig:control-a}
    \end{subfigure}
    \begin{subfigure}[t]{1.\linewidth}
      \centering\includegraphics[width=\linewidth]{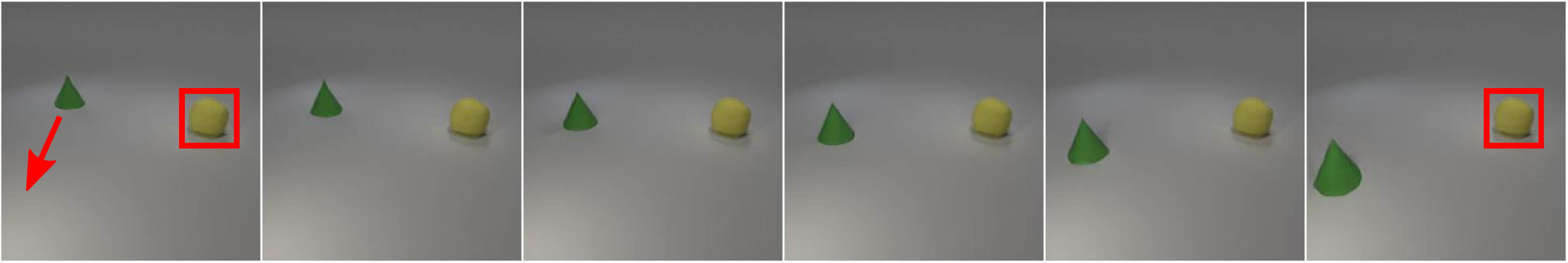}
      \caption{Translation of Left Object (Ours)}
      \label{fig:control-b} 
    \end{subfigure}
    \begin{subfigure}[t]{.493\linewidth}
      \centering\includegraphics[width=1.\linewidth]{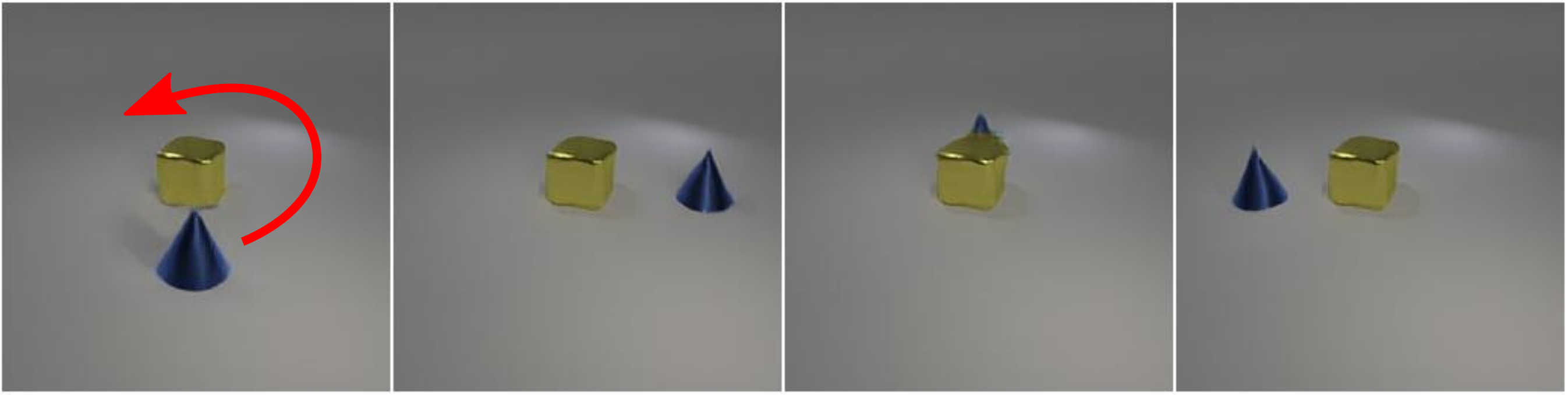}
      \caption{Circular Translation (Ours)}
      \label{fig:control-c}
    \end{subfigure}
    \begin{subfigure}[t]{.493\linewidth}
      \centering\includegraphics[width=1.\linewidth]{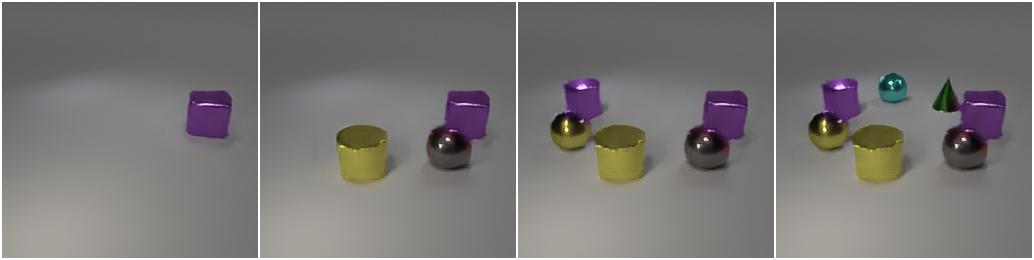}
      \caption{Add Objects (Ours)}
      \label{fig:control-d} 
    \end{subfigure}
    \vspaceunderfig\vspace{-0.1cm}
  \caption{
      \textbf{Controllable Image Generation.}
      While most generative models operate in 2D, we incorporate a compositional 3D scene representation into the generative model.
      This leads to more consistent image synthesis results, \eg note how, in contrast to our method, translating one object might change the other when operating in 2D (Fig.\ \ref{fig:control-a} and \ref{fig:control-b}).
      It further allows us to perform complex operations like circular translations (\figref{fig:control-c}) or adding more objects at test time (\figref{fig:control-d}).
      Both methods are trained unsupervised on raw unposed image collections of two-object scenes.
    }
  \label{fig:controllable-generation}
\end{figure}
The ability to generate and manipulate photorealistic image content is a long-standing goal of computer vision and graphics.
Modern computer graphics techniques achieve impressive results and are industry standard in gaming and movie productions. However, they are very hardware expensive and require substantial human labor for 3D content creation and arrangement.

In recent years, the computer vision community has made great strides towards highly-realistic image generation. In particular, Generative Adversarial Networks (GANs)~\cite{Goodfellow2014NIPS} emerged as a powerful class of generative models.
They are able to synthesize photorealistic images at resolutions of $1024^2$ pixels and beyond~\cite{Karras2019CVPR, Karras2020CVPRi, Choi2018CVPR, Choi2020CVPR, Brock2019ICLR}. 

Despite these successes, synthesizing realistic 2D images is not the only aspect required in applications of generative models.
The generation process should also be controllable in a simple and consistent manner.
To this end, many works~\cite{Chen2016NIPS, Lee2020ARXIV, Peebles2020ECCV, Reed2014ICML, Karras2019CVPR, Zhao2018ECCV, Kwak2016ARXIV, Locatello2020NEURIPS, Liu2020ARXIV, Goyal2019ARXIV, Zhu2015ICCV} investigate how disentangled representations can be learned from data without explicit supervision.
Definitions of disentanglement vary~\cite{Bengio2013PAMI,Locatello2019ICML}, but commonly refer to being able to control an attribute of interest, \eg object shape, size, or pose, without changing other attributes.
Most approaches, however, do not consider the compositional nature of scenes and operate in the 2D domain, ignoring that our world is three-dimensional.
This often leads to entangled representations~(\figref{fig:controllable-generation}) and control mechanisms are not built-in, but need to be discovered in the latent space a~posteriori.
These properties, however, are crucial for successful applications, \eg a movie production where complex object trajectories need to be generated in a consistent manner.

Several recent works therefore investigate how to incorporate 3D representations, such as voxels~\cite{Henzler2019ICCV, Nguyen-Phuoc2019ICCV, Nguyen-Phuoc2020NEURIPSp}, primitives~\cite{Liao2020CVPR}, or radiance fields~\cite{Schwarz2020NEURIPS}, directly into generative models.
While these methods allow for impressive results with built-in control, they are mostly restricted to single-object scenes and results are less consistent for higher resolutions and more complex and realistic imagery (\eg scenes with objects not in the center or cluttered backgrounds).

\vspace{-0.02cm}\boldparagraph{Contribution} In this work, we introduce \textit{GIRAFFE}, a novel method for generating scenes in a controllable and photorealistic manner while training from raw unstructured image collections.
Our key insight is twofold:
First, incorporating a compositional 3D scene representation directly into the generative model leads to more controllable image synthesis.
Second, combining this explicit 3D representation with a neural rendering pipeline results in faster inference and more realistic images.
To this end, we represent scenes as compositional generative neural feature fields (\figref{fig:teaser}).
We volume render the scene to a feature image of relatively low resolution to save time and computation.
A neural renderer processes these feature images and outputs the final renderings.
This way, our approach achieves high-quality images and scales to real-world scenes.
We find that our method allows for controllable image synthesis of single-object as well as multi-object scenes when trained on raw unstructured image collections.
Code and data is available at \url{https://github.com/autonomousvision/giraffe}.
\section{Related Work}
\label{sec:rel-work}

\vspace{-0.13cm}\boldparagraphnovspace{GAN-based Image Synthesis}
Generative Adversarial Networks (GANs)~\cite{Goodfellow2014NIPS} have been shown to allow for photorealistic image synthesis at resolutions of $1024^2$ pixels and beyond~\cite{Karras2019CVPR, Karras2020CVPRi, Choi2018CVPR, Choi2020CVPR, Brock2019ICLR}.
To gain better control over the synthesis process, many works investigate how factors of variation can be disentangled without explicit supervision.
They either modify the training objective~\cite{Peebles2020ECCV, Karras2020CVPRi, Chen2016NIPS} or network architecture~\cite{Karras2019CVPR}, or 
investigate latent spaces of well-engineered and pre-trained generative models~
\cite{Harkonen2020ARXIV, Collins2020CVPR, Jahanian2020ICLR, Shen2020CVPR, Goetschalckx2019ICCV, Abdal2020ARXIV, Zhan2020ARXIV}.
All of these works, however, do not explicitly model the compositional nature of scenes.
Recent works therefore investigate how the synthesis process can be controlled at the object-level~\cite{Engelcke2020ICLRi, Yang2017ICLR, Ehrhardt2020ARXIV, Arandjelovic2019ARXIV, Burgess2019ARXIV, Greff2019ICML, Steenkiste2020NN, Anciukevicius2020ARXIV, Li2020NEURIPS}.
While achieving photorealistic results, all aforementioned works model the image formation process in 2D, ignoring the three-dimensional structure of our world.
In this work, we advocate to model the formation process directly in 3D for better disentanglement and more controllable synthesis.

\boldparagraph{Implicit Functions}
Using implicit functions to represent 3D geometry has gained popularity in learning-based 3D reconstruction~\cite{Mescheder2019CVPR, Park2019CVPR, Chen2019CVPR,Oechsle2019ICCV, Saito2019ICCV,Niemeyer2019ICCV,Genova2019ICCV,Chen2019ICCV, Michalkiewicz2019ICCV} 
and has been extended to scene-level reconstruction~\cite{Chabra2020ECCV, Jigan2020CVPR, Peng2020ECCV, sitzmann2020NEURIPS,Chibane2020NEURIPS}.
To overcome the need of 3D supervision, several works~\cite{Liu2019NEURIPS,Niemeyer2020CVPR,Sitzmann2019NIPS,Yariv2020NEURIPS, Liu2020CVPR} propose differentiable rendering techniques.
Mildenhall~\etal~\cite{Mildenhall2020ECCV} propose Neural Radiance Fields (NeRFs) in which they combine an implicit neural model with volume rendering for novel view synthesis of complex scenes.
Due to their expressiveness, we use a generative variant of NeRFs as our object-level representation.
In contrast to our method, the discussed works require multi-view images with camera poses as supervision, train a single network per scene, and are not able to generate novel scenes.
Instead, we learn a generative model from unstructured image collections which allows for controllable, photorealistic image synthesis of generated scenes. 

\boldparagraph{3D-Aware Image Synthesis}
Several works investigate how 3D representations can be incorporated as inductive bias into generative models~\cite{Henzler2019ICCV, Lunz2020ARXIV, Nguyen-Phuoc2019ICCV, Nguyen-Phuoc2020NEURIPSp, Liao2020CVPR, Schwarz2020NEURIPS, Gadelha2017THREEDV, Rezende2016NIPS, Henderson2020ARXIV, Henderson2020CVPR, Henderson2019IJCV}.
While many approaches use additional supervision~\cite{Zhu2018NIPS, Wang2016ECCV, Alhaija2018ACCV, Chen2020ARXIV, Wu2016NIPS}, we focus on works which are trained on raw image collections like our approach. \\
Henzler~\etal~\cite{Henzler2019ICCV} learn voxel-based representations using differentiable rendering.
The results are 3D controllable, but show artifacts due to the limited voxel resolutions caused by their cubic memory growth. 
Nguyen-Phuoc~\etal~\cite{Nguyen-Phuoc2019ICCV, Nguyen-Phuoc2020NEURIPSp} propose voxelized feature-grid representations which are rendered to 2D via a reshaping operation.
While achieving impressive results, training becomes less stable and results less consistent for higher resolutions. %
Liao~\etal~\cite{Liao2020CVPR} use abstract features in combination with primitives and differentiable rendering.
While handling multi-object scenes, they require additional supervision in the form of pure background images which are hard to obtain for real-world scenes.
Schwarz~\etal~\cite{Schwarz2020NEURIPS} propose Generative Neural Radiances Fields (GRAF).
While achieving controllable image synthesis at high resolutions, this representation is restricted to single-object scenes and results degrade on more complex, real-world imagery.
In contrast, we incorporate compositional 3D scene structure into the generative model such that it naturally handles multi-object scenes.
Further, by integrating a neural rendering pipeline~\cite{Eslami2018Science, Nguyen-Phuoc2018NIPS, Sitzmann2019NIPS, Kato2018CVPR, Liu2019ICCV, Thies2019TOG, Tewari2020CGF, Sitzmann2019CVPR, Kato2020ARXIV}, our model scales to more complex, real-world data. %

\section{Method}
\begin{figure*}
    \centering
    \includegraphics[width=\linewidth]{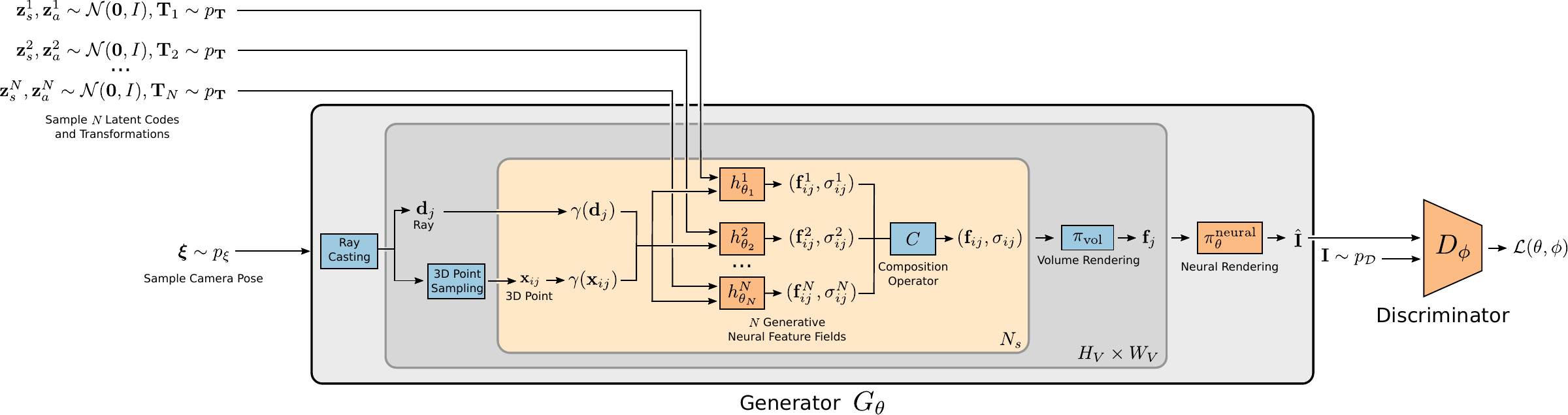}
    \vspaceunderfig
    \vspace{-.35cm}
    \caption{
        \textbf{GIRAFFE.}
        Our generator $G_\theta$ takes a camera pose $\bxi$ and $N$ shape and appearance codes $\zshape^i, \zapp^i$ and affine transformations $\bT_i$ as input and synthesizes an image of the generated scene which consists of $N-1$ objects and a background.
        The discriminator $D_\phi$ takes the generated image $\igen$ and the real image $\bI$ as input and our full model is trained with an adversarial loss.
        At test time, we can control the camera pose, the shape and appearance codes of the objects, and the objects' poses in the scene.
        Orange indicates learnable and blue non-learnable operations.
        }
    \label{fig:method-overview}
\end{figure*}
Our goal is a controllable image synthesis pipeline which can be trained from raw image collections without additional supervision.
In the following, we discuss the main components of our method.
First, we model individual objects as neural feature fields (\secref{subsec:objects}).
Next, we exploit the additive property of feature fields to composite scenes from multiple individual objects (\secref{subsec:scenes}).
For rendering, we explore an efficient combination of volume and neural rendering techniques (\secref{subsec:rendering}).
Finally, we discuss how we train our model from raw image collections (\secref{subsec:training}).
\figref{fig:method-overview} contains an overview of our method.

\subsection{Objects as Neural Feature Fields}
\label{subsec:objects}
\vspace{-0.13cm}\boldparagraphnovspace{Neural Radiance Fields} A radiance field is a continuous function $f$ which maps a 3D point $\mathbf{x} \in \mathbb{R}^3$ and a viewing direction $\mathbf{d} \in \mathbb{S}^2$ to a volume density $\sigma \in \mathbb{R}^+$ and an RGB color value $\mathbf{c} \in \mathbb{R}^3$. 
 A key observation in~\cite{Mildenhall2020ECCV, Tancik2020NEURIPS} is that the low dimensional input $\mathbf{x}$ and $\mathbf{d}$ needs to be mapped to higher-dimensional features to be able to represent complex signals when $f$ is parameterized with a neural network. More specifically, a pre-defined positional encoding is applied element-wise to each component of $\mathbf{x}$ and $\mathbf{d}$:
 \begin{align}\begin{split}\label{eq:pos-enc}
     \gamma(t, L) &= \\ (\sin(2^0 t \pi), &\cos(2^0 t \pi), \dots, \sin(2^L t \pi), \cos(2^L t \pi))
\end{split}\end{align}
where $t$ is a scalar input, \eg a component of $\bx$ or $\bd$, and $L$ the number of frequency octaves.
In the context of generative models, we observe an additional benefit of this representation: It introduces an inductive bias to learn 3D shape representations in canonical orientations which otherwise would be arbitrary (see~\figref{fig:canonical-pose}).

Following implicit shape representations~\cite{Mescheder2019CVPR, Chen2019CVPR, Park2019CVPR}, 
 Mildenhall~\etal~\cite{Mildenhall2020ECCV} propose to learn Neural Radiance Fields (NeRFs) by parameterizing $f$ with a multi-layer perceptron (MLP):
\begin{align}\begin{split}
    f_\theta: \mathbb{R}^{L_\mathbf{x}} \times \mathbb{R}^{L_\mathbf{d}} &\to \mathbb{R}^+ \times \mathbb{R}^3 \\ (\gamma(\mathbf{x}), \gamma(\mathbf{d})) &\mapsto (\sigma, \mathbf{c})
\end{split}\end{align}
where $\theta$ indicate the network parameters and $L_\bx, L_\bd$ the output dimensionalities of the positional encodings.

\boldparagraph{Generative Neural Feature Fields}
 While \cite{Mildenhall2020ECCV}~fits $\theta$ to multiple posed images of a single scene, Schwarz~\etal~\cite{Schwarz2020NEURIPS} propose a generative model for Neural Radiance Fields (GRAF) that is trained from unposed image collections.
To learn a latent space of NeRFs, they condition the MLP on shape and appearance codes $\zshape, \zapp \sim \mathcal{N}(\mathbf{0}, I)$:
\begin{align}\begin{split}
    g_\theta: \mathbb{R}^{L_\mathbf{x}} \times \mathbb{R}^{L_\mathbf{d}} \times \mathbb{R}^{M_s} \times \mathbb{R}^{M_a} &\to \mathbb{R}^+ \times \mathbb{R}^3 \\ (\gamma(\mathbf{x}), \gamma(\mathbf{d}), \zshape, \zapp) &\mapsto (\sigma, \mathbf{c})
\end{split}\end{align}
where $M_s, M_a$ are the dimensionalities of the latent codes.

In this work we explore a more efficient combination of volume and neural rendering.
We replace GRAF's formulation for the three-dimensional color output $\bc$ with a more generic $M_f$-dimensional feature $\mathbf{f}$ and represent objects as Generative Neural Feature Fields:
\begin{align}\begin{split}
    h_\theta: \mathbb{R}^{L_\mathbf{x}} \times \mathbb{R}^{L_\mathbf{d}} \times \mathbb{R}^{M_s} \times \mathbb{R}^{M_a} &\to \mathbb{R}^+ \times \mathbb{R}^{M_f} \\ (\gamma(\mathbf{x}), \gamma(\mathbf{d}), \zshape, \zapp) &\mapsto (\sigma, \mathbf{f})
\end{split}\end{align}

\boldparagraph{Object Representation}
A key limitation of NeRF and GRAF is that the entire scene is represented by a single model.
As we are interested in disentangling different entities in the scene, we need control over the pose, shape and appearance of \textit{individual} objects (we consider the background as an object as well).
We therefore represent each object using a separate feature field in combination with an affine transformation %
\begin{align}\begin{split}\label{eq:object-transformation}
    \bT = \{\mathbf{s}, \mathbf{t}, \mathbf{R}\} 
\end{split}\end{align}
where $\mathbf{s}, \mathbf{t} \in \mathbb{R}^3$ indicate scale and translation parameters, and $\mathbf{R} \in SO(3)$ a rotation matrix.
Using this representation, we transform points from object to scene space as follows:
\begin{align}
    \begin{split}
    k(\bx) = \bR \cdot   \begin{bmatrix}
        s_{1} & & \\
        & s_{2} & \\
        & & s_{3}
      \end{bmatrix} \cdot \bx + \bt
    \end{split}
\end{align}
In practice, we volume render in scene space and evaluate the feature field in its canonical~object~space (see~\figref{fig:teaser}):
\begin{align}
    \begin{split}
        (\sigma, \mathbf{f}) = h_{\theta}(\gamma(k^{-1}(\mathbf{x})), \gamma(k^{-1}(\mathbf{d})), \zshape, \zapp)
    \end{split}
\end{align} 
This allows us to arrange multiple objects in a scene. 
All object feature fields share their weights and $\bT$ is sampled from a dataset-dependent distribution (see~\secref{subsec:training}).

\subsection{Scene Compositions} 
\label{subsec:scenes}\vspace{-0.12cm}
As discussed above, we describe scenes as compositions of $N$ entities where the first $N - 1$ are the objects in the scene and the last represents the background.
We consider two cases: First, $N$ is fixed across the dataset such that the images always contain $N-1$ objects plus the background.
Second, $N$ is varied across the dataset.
In practice, we use the same representation for the background as for objects except that we fix the scale and translation parameters $\mathbf{s}_{N}, \mathbf{t}_{N}$ to span the entire scene, and to be centered at the scene space origin.

\boldparagraph{Composition Operator} To define the composition operator $C$, let's recall that a feature field of a single entity $h_{\theta_i}^i$ predicts a density $\sigma_i \in \mathbb{R}^+$ and a feature vector $\mathbf{f}_i \in \mathbb{R}^{M_f}$ for a given point $\mathbf{x}$ and viewing direction $\mathbf{d}$.
When combining non-solid objects, a natural choice~\cite{Drebin1988SIGGRAPH} for the overall density at $\mathbf{x}$ is to sum up the individual densities and to use the density-weighted mean to combine all features at $(\mathbf{x}, \mathbf{d})$:
\begin{align}\label{eq:composition}
    \begin{split}
        \comp(\mathbf{x}, \mathbf{d}) = \left( \sigma, \frac{1}{\sigma } \sum_{i=1}^{N} \sigma_i \mathbf{f}_i \right)
        \text{, where}\quad \sigma = \sum_{i=1}^{N} \sigma_i
    \end{split}
\end{align}
While being simple and intuitive, this choice for $C$ has an additional benefit: We ensure gradient flow to all entities with a density greater than $0$.

\subsection{Scene Rendering}
\label{subsec:rendering}
\vspace{-0.13cm}\boldparagraphnovspace{3D Volume Rendering}
While previous works~\cite{Mildenhall2020ECCV, Schwarz2020NEURIPS, Liu2020NEURIPS, Brualla2020ARXIV} volume render an RGB color value, we extend this formulation to rendering an $M_f$-dimensional feature vector $\mathbf{f}$.

For given camera extrinsics $\bxi$, let $\{\mathbf{x}_j\}_{j=1}^{N_s}$ be sample points along the camera ray $\mathbf{d}$ for a given pixel, and $(\sigma_j, \mathbf{f}_j)=\comp(\bx_j, \bd)$ the corresponding densities and feature vectors of the field.
The volume rendering operator $\rv$~\cite{Kajiya1984SIGGRAPH} maps these evaluations to the pixel's final feature vector $\mathbf{f}$:
\begin{align}
    \begin{split}\label{eq:rendering-operator}
        \rv: (\mathbb{R}^+ \times \mathbb{R}^{M_f})^{N_s} \to \mathbb{R}^{M_f}, \quad \{\sigma_j, \mathbf{f}_j\}_{j=1}^{N_s} \mapsto \mathbf{f}
    \end{split}
\end{align}
Using numerical integration as in~\cite{Mildenhall2020ECCV}, $\mathbf{f}$ is obtained as 
\begin{align}
    \begin{split}\label{eq:render-final-c}
        \mathbf{f} = \sum_{j=1}^{N_s} \tau_j \alpha_j \mathbf{f}_j \quad
        \tau_j = \prod_{k=1}^{j-1} (1 - \alpha_k) \quad \alpha_j = 1 - e^{-\sigma_j\delta_j}
    \end{split}
\end{align}
where $\tau_j$ is the transmittance, $\alpha_j$ the alpha value for $\mathbf{x}_j$, and $\delta_j = {\left|\left|\mathbf{x}_{j+1} - \mathbf{x}_j \right|\right|}_2$ the distance between neighboring sample points.
The entire feature image is obtained by evaluating $\rv$ at every pixel.
For efficiency, we render feature images at resolution $16^2$ which is lower than the output resolution of $64^2$ or $256^2$ pixels.
We then upsample the low-resolution feature maps to higher-resolution RGB images using 2D neural rendering.
As evidenced by our experiments, this has two advantages: increased rendering speed and improved image quality.

\boldparagraph{2D Neural Rendering}
\begin{figure}
    \centering
    \includegraphics[width=\linewidth]{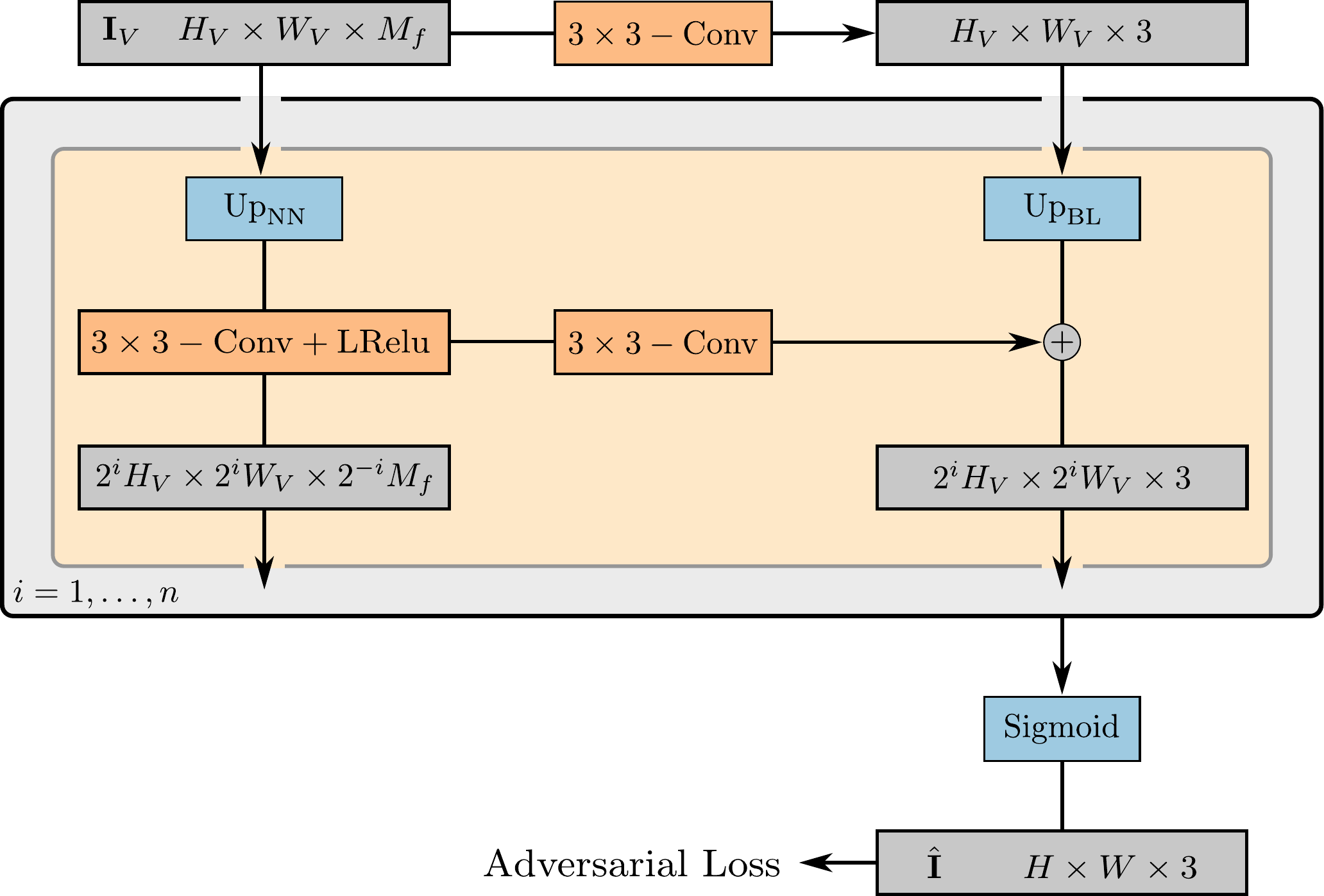}
    \vspaceunderfig\vspace{-.2cm}
    \caption{
        \textbf{Neural Rendering Operator.} 
            The feature image $\mathbf{I}_V$ is processed by $n$ blocks of nearest neighbor upsampling and $3 \times 3$ convolutions with leaky ReLU activations.
            At every resolution, we map the feature image to an RGB image with a $3 \times 3$ convolution and add it to the previous output via bilinear upsampling.
            We apply a sigmoid activation to obtain the final image $\igen$.
            Gray color indicates outputs, orange learnable, and blue non-learnable operations.
        }
    \label{fig:neural-rendering}
\end{figure}
The neural rendering operator
\begin{align} 
    \begin{split}
        \rn: \mathbb{R}^{H_V \times W_V \times M_f} \to \mathbb{R}^{H \times W \times 3}
    \end{split}
\end{align}
with weights $\theta$ maps the feature image $\mathbf{I}_V \in \mathbb{R}^{H_V \times W_V \times M_f}$ to the final synthesized image $\igen \in \mathbb{R}^{H \times W \times 3}$.
We parameterize $\rn$ as a 2D convolutional neural network (CNN) with leaky ReLU~\cite{XU2015ARXIV, Maas2013ICMLWORK} activation (\figref{fig:neural-rendering}) and combine nearest neighbor upsampling with $3\times3$ convolutions to increase the spatial resolution. We choose small kernel sizes and no intermediate layers to only allow for spatially small refinements to avoid entangling global scene properties during image synthesis while at the same time allowing for increased output resolutions.
Inspired by~\cite{Karras2020CVPRi}, we map the feature image to an RGB image at every spatial resolution, and add the previous output to the next via bilinear upsampling. 
These skip connections ensure a strong gradient flow to the feature fields.
We obtain our final image prediction $\igen$ by applying a sigmoid activation to the last RGB layer.
We validate our design choices in an ablation study (\tabref{tab:ablation-study}).

\subsection{Training}
\label{subsec:training}
\vspace{-.13cm}\boldparagraphnovspace{Generator} We denote the full generative process formally as %
\begin{align}
    \begin{split}
        G_\theta (\{\zshape^i, \zapp^i, \ti\}_{i=1}^{N}, \bxi) = \rn(\mathbf{I}_V)& \\
        \text{where} \quad \mathbf{I}_V = \{ \rv ( \{ \comp(\mathbf{x}_{jk}, \mathbf{d}_k) \}_{j=1}^{N_s} ) &\}_{k=1}^{H_V \times W_V}
    \end{split}
\end{align}
and $N$ is the number of entities in the scene, $N_s$ the number of sample points along each ray, $\mathbf{d}_k$ is the ray for the $k$-th pixel, and $\mathbf{x}_{jk}$ the $j$-th sample point for the $k$-th pixel / ray.

\boldparagraph{Discriminator} We parameterize the discriminator $D_\phi$ as a CNN~\cite{Radford2016ICLR} with leaky~ReLU~activation.

\boldparagraph{Training} During training, we sample the the number of entities in the scene $N \sim p_N$, the latent codes $\zshape^i, \zapp^i \sim \mathcal{N}(\mathbf{0}, I)$, as well as a camera pose $\bxi \sim p_{\xi}$ and object-level transformations $\ti \sim p_{T}$.
In practice, we define $p_{\xi}$ and $p_{T}$ as uniform distributions over dataset-dependent camera elevation angles and valid object transformations, respectively.\footnote{Details can be found in the supplementary material.}
The motivation for this choice is that in most real-world scenes, objects are arbitrarily rotated, but not tilted due to gravity.
The observer (the camera in our case), in contrast, can freely change its elevation angle \wrt the scene.

We train our model with the non-saturating GAN objective~\cite{Goodfellow2014NIPS} and $R_1$ gradient penalty~\cite{Mescheder2018ICML}
\begin{equation}
    \begin{split}
    &\cV(\theta, \phi) =\\
    &\nE_{\bz_s^i, \bz_a^i\sim \mathcal{N},\,\bxi\sim p_{\xi},\,\ti\sim p_{T}}
    \left[
    f(D_\phi(G_\theta(\{ \zshape^i, \zapp^i, \ti\}_i, \bxi))\right]\\
    &+ \nE_{\bI \sim p_{\cD}}\left[f(-D_\phi(\bI))-\, \lambda {\Vert \nabla D_\phi(\bI)\Vert}^2 \right] 
    \end{split}
\end{equation}
where $f(t)=-\log(1+\exp(-t))$, $\lambda = 10$, and $p_\cD$ indicates the data distribution.

\subsection{Implementation Details}
\label{subsec:implementation-details}\vspace{-0.12cm}
All object feature fields $\{h_{\theta_i}^i\}_{i=1}^{N-1}$ share their weights and we parametrize them as MLPs with ReLU~activations.
We use $8$ layers with a hidden dimension of $128$ and a density and a feature head of dimensionality $1$ and $M_f=128$, respectively.
For the background feature field $h_{\theta_N}^N$, we use half the layers and hidden dimension.
We use $L_\mathbf{x} = 2 \cdot 3 \cdot 10$ and $L_\mathbf{d} = 2 \cdot 3 \cdot 4$ for the positional encodings.
We sample $M_s = 64$ points along each ray and render the feature image $\mathbf{I}_V$ at $16^2$ pixels. 
We use an exponential moving average~\cite{Yazici2019ICLR} with decay $0.999$ for the weights of the generator.
We use the RMSprop optimizer~\cite{Tieleman2012Coursera} with a batch size of $32$ and learning rates of \num{1e-4} and \num{5e-4} for the discriminator and generator, respectively.
For experiments at $256^2$ pixels, we set $M_f=256$ and half the generator learning rate to \num{2.5e-4}.

\section{Experiments}

\vspace{-.13cm}\boldparagraphnovspace{Datasets} We report results on commonly-used single-object datasets \textit{Chairs}~\cite{Oechsle2020THREEDV}, \textit{Cats}~\cite{Zhang2003ICCV}, \textit{CelebA}~\cite{Liu2015ICCV}, and \textit{CelebA-HQ}~\cite{Karras2018ICLR}.
The first consists of synthetic renderings of Photoshape~chairs~\cite{Park2018ACM}, and the others are image collections of cat and human faces, respectively.
The data complexity is limited as the background is purely white or only takes up a small part of the image. 
We further report results on the more challenging single-object, real-world datasets \textit{CompCars}~\cite{Yang2015CVPR}, \textit{LSUN Churches}~\cite{Yu2015ARXIV}, and \textit{FFHQ}~\cite{Karras2019CVPR}.
For \textit{CompCars}, we randomly crop the images to achieve more variety of the object's position in the image.\footnote{\noindent We do not apply random cropping for \cite{Henzler2019ICCV} and \cite{Schwarz2020NEURIPS} as we find that they cannot handle scenes with non-centered objects (see supplementary).}
For these datasets, disentangling objects is more complex as the object is not always in the center and the background is more cluttered and takes up a larger part of the image.
To test our model on multi-object scenes, we use the script from~\cite{Johnson2017CVPR} to render scenes with $2$, $3$, $4$, or $5$ random primitives (\textit{Clevr-N}).
To test our model on scenes with a varying number of objects, we also run our model on the union of them ($\textit{Clevr-2345}$).

\boldparagraph{Baselines} We compare against voxel-based PlatonicGAN~\cite{Henzler2019ICCV}, BlockGAN~\cite{Nguyen-Phuoc2020NEURIPSp}, and HoloGAN~\cite{Nguyen-Phuoc2019ICCV}, and radiance field-based GRAF~\cite{Schwarz2020NEURIPS} (see~\secref{sec:rel-work} for a discussion of the methods).
We further compare against HoloGAN w/o 3D~Conv, a variant of~\cite{Nguyen-Phuoc2019ICCV} proposed in~\cite{Schwarz2020NEURIPS} for higher resolutions.
We additionally report a ResNet-based~\cite{He2016CVPR} 2D~GAN~\cite{Mescheder2018ICML} for reference.

\boldparagraph{Metrics} We report the Frechet Inception Distance (FID) score~\cite{Heusel2017NIPS} to quantify image quality.
We use \num{20000} real and fake samples to calculate the FID score.

\subsection{Controllable Scene Generation}
\begin{figure}
  \centering
  \begin{tabular}{cccccc}
    \includegraphics[width=.7cm]{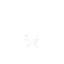} & \includegraphics[width=.7cm]{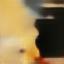} & \includegraphics[width=.7cm]{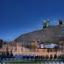} & \includegraphics[width=.7cm]{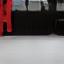} &
    \includegraphics[width=.7cm]{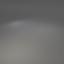} & \includegraphics[width=.7cm]{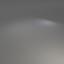}  \\
    \includegraphics[width=.7cm]{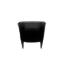} & \includegraphics[width=.7cm]{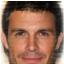} & \includegraphics[width=.7cm]{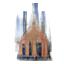} & 
    \includegraphics[width=.7cm]{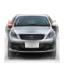} &
\includegraphics[width=.7cm]{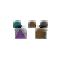} & \includegraphics[width=.7cm]{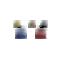}  \\
    \includegraphics[width=.7cm]{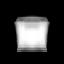} & \includegraphics[width=.7cm]{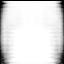} & \includegraphics[width=.7cm]{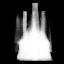} & 
    \includegraphics[width=.7cm]{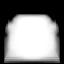} &
 \includegraphics[width=.7cm]{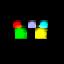} & \includegraphics[width=.7cm]{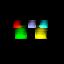}  \\
    \includegraphics[width=.7cm]{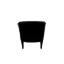} & \includegraphics[width=.7cm]{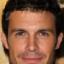} & \includegraphics[width=.7cm]{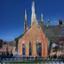} & 
    \includegraphics[width=.7cm]{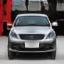} &
\includegraphics[width=.7cm]{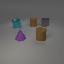} & \includegraphics[width=.7cm]{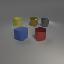}  \\
    \bottomrule
    {\footnotesize Chairs} & {\footnotesize CelebA} & {\footnotesize Churches} & {\footnotesize Cars} & {\footnotesize Clevr-5} & {\footnotesize Clevr-2345}
\end{tabular}

  \vspaceunderfig%
  \caption{
      \textbf{Scene Disentanglement.}
      From top to bottom, we show only backgrounds, only objects, color-coded object alpha maps, and the final synthesized images at $64^2$ pixel resolution.
      Disentanglement emerges without supervision, and the model learns to generate plausible backgrounds although the training data only contains images with objects.
      }
  \label{fig:disentanglement}
\end{figure}
\begin{figure}
  \includegraphics[width=1.\linewidth]{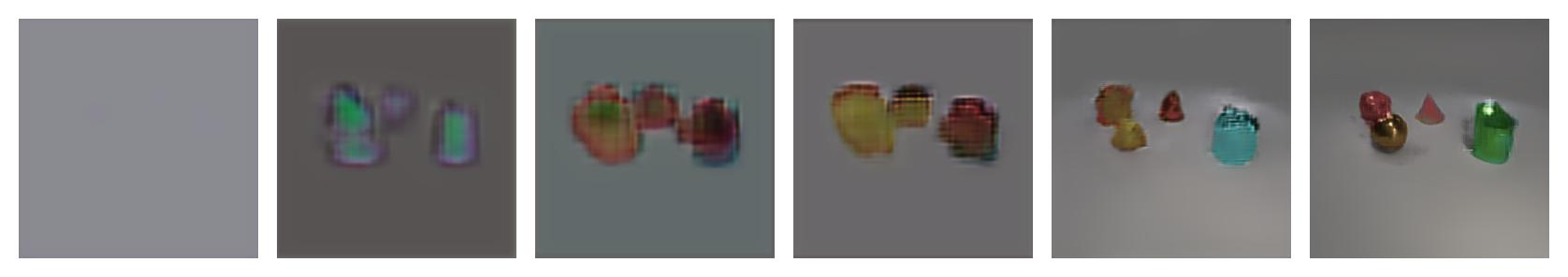}
  \vspaceunderfigext%
  \caption{
      \textbf{Training Progression.}
          We show renderings of our model on \textit{Clevr-2345} at $256^2$ pixels after $0$, $1$, $2$, $3$, $10$, and $100$-thousand iterations.
          Unsupervised disentanglement emerges already at the very beginning of training.
      }
  \label{fig:training-progression}
\end{figure}
\begin{figure*}
  \centering
  \begin{subfigure}[t]{.495\linewidth}
      \centering\includegraphics[width=\linewidth]{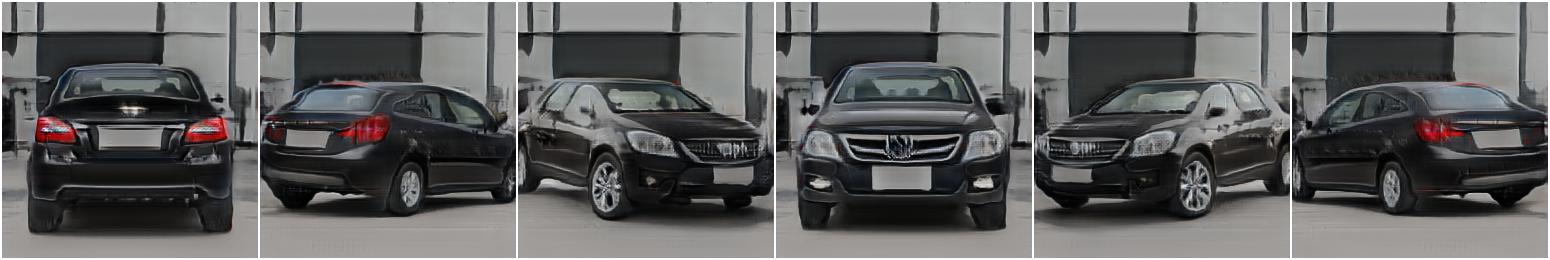}\vspace{-.1cm}
      \subcaption{Object Rotation}
      \label{subfig:control-elevation}
  \end{subfigure}
  \begin{subfigure}[t]{0.495\linewidth}
      \centering\includegraphics[width=\linewidth]{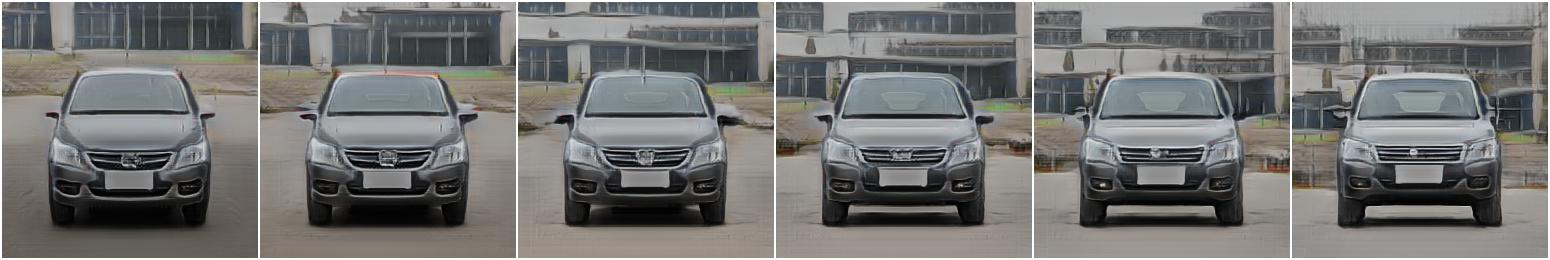}\vspace{-.1cm}
      \subcaption{Camera Elevation}
      \label{subfig:control-rotation}
    \end{subfigure}
    \begin{subfigure}[t]{1.\linewidth}
        \centering\includegraphics[width=0.495\linewidth]{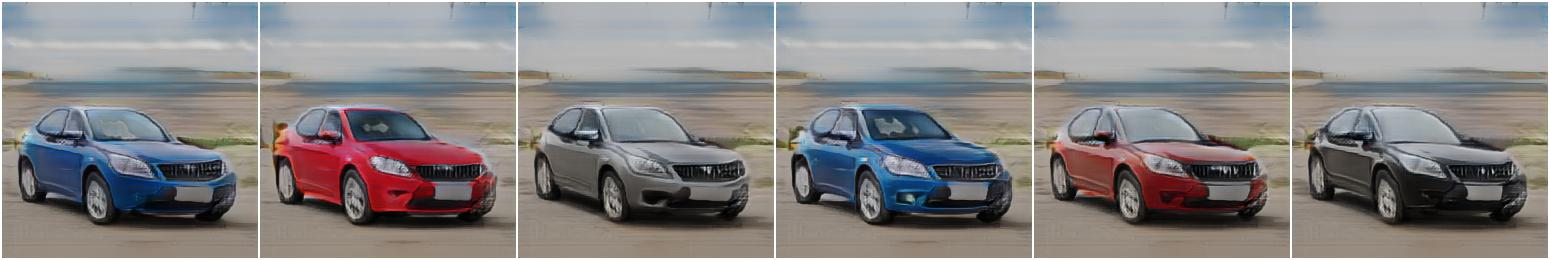}
        \includegraphics[width=0.495\linewidth]{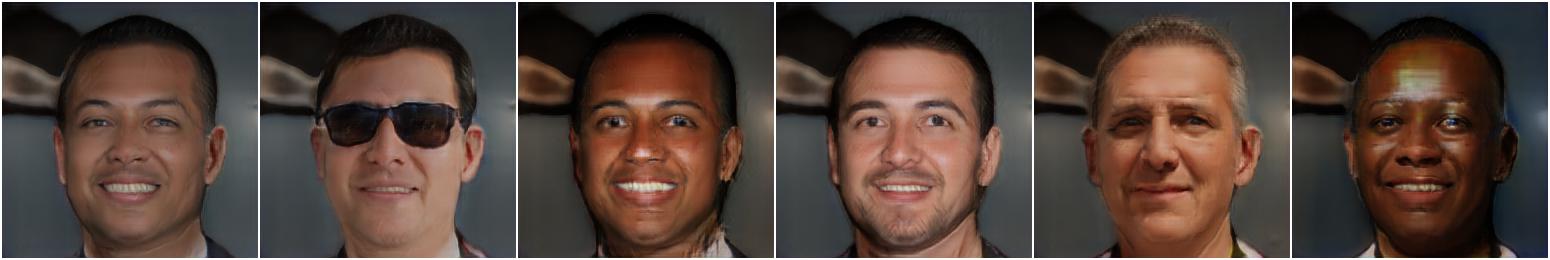}
        \label{subfig:control-appearance}\vspace{-.1cm}
        \subcaption{Object Appearance}
      \end{subfigure}
      \begin{subfigure}[t]{.495\linewidth}
          \centering\includegraphics[width=\linewidth]{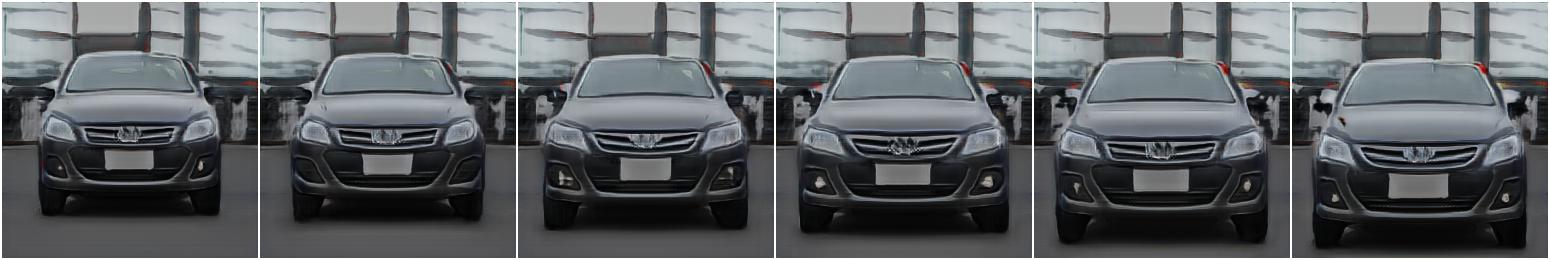}\vspace{-.1cm}
          \subcaption{Depth Translation}
          \end{subfigure}
        \begin{subfigure}[t]{.495\linewidth}
            \centering\includegraphics[width=\linewidth]{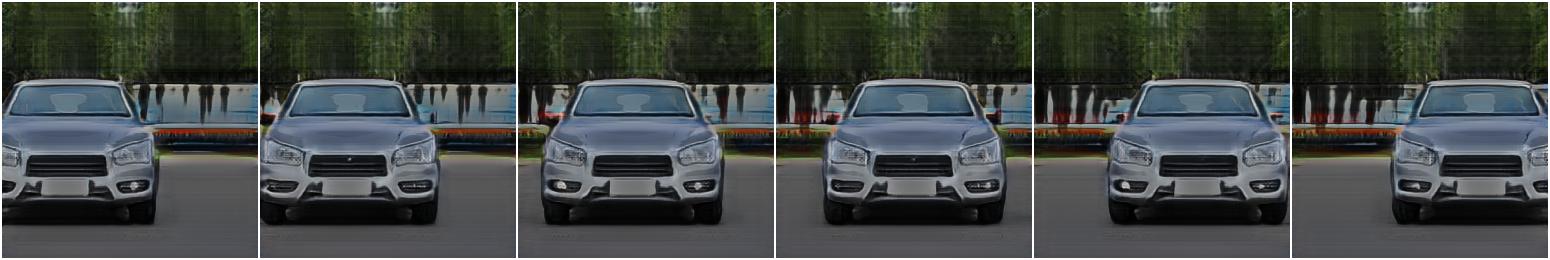}\vspace{-.1cm}
            \subcaption{Horizontal Translation}
           \end{subfigure}
        \begin{subfigure}[t]{1.\linewidth}
            \centering\includegraphics[width=\linewidth]{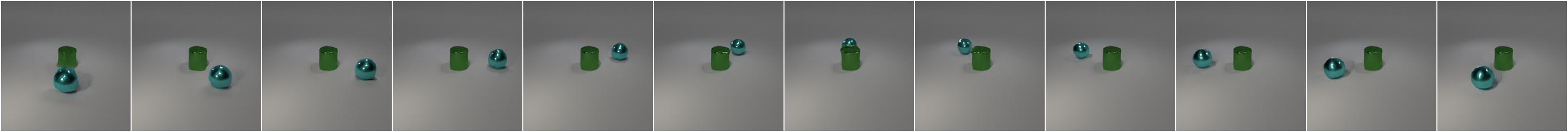}\vspace{-.1cm}
            \subcaption{Circular Translation of One Object Around Another Object}
            \label{subfig:control-circle}
          \end{subfigure}
  \vspaceunderfig\vspace{-.1cm}
  \caption{
      \textbf{Controllable Scene Generation at $256^2$ Pixel Resolution.} 
      Controlling the generated scenes during image synthesis: Here we rotate or translate objects, change their appearances, and perform complex operations like circular translations.
      }
      \vspace{-.3cm}
  \label{fig:controllable-scene-gen}
\end{figure*}
\begin{figure} 
  \centering
  \begin{subfigure}[t]{1.\linewidth}
  \includegraphics[width=1.\linewidth]{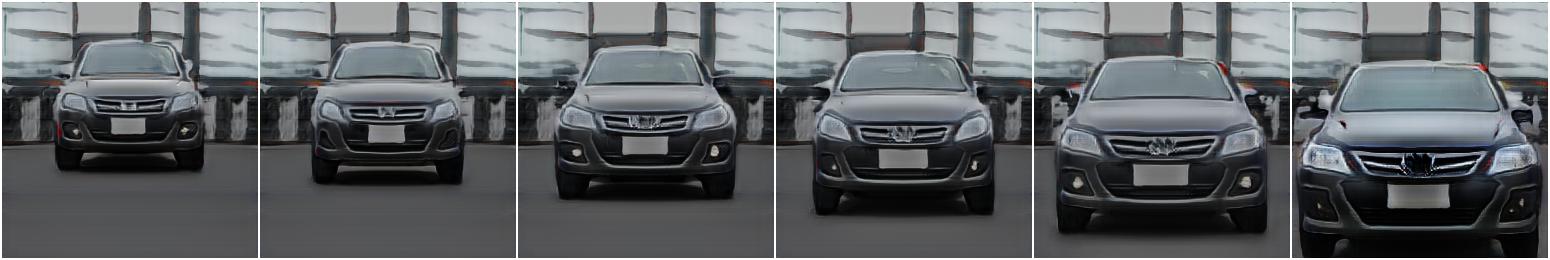}\vspace{-.1cm}
    \subcaption{Increase Depth Translation}
    \label{subfig:increase-z-translation}
\end{subfigure}
  \begin{subfigure}[t]{1.\linewidth}
  \includegraphics[width=1.\linewidth]{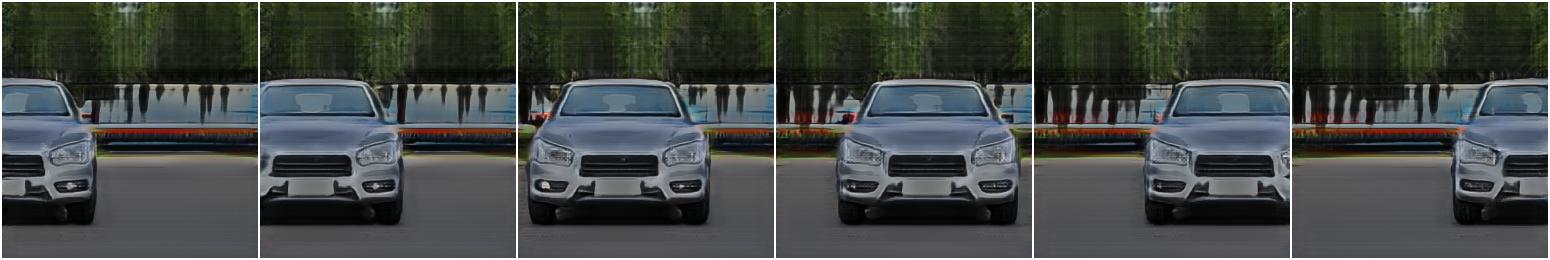}\vspace{-.1cm}
    \subcaption{Increase Horizontal Translation}
    \label{subfig:increase-z-translation}
\end{subfigure}
  \begin{subfigure}[t]{1.\linewidth}
  \includegraphics[width=1.\linewidth]{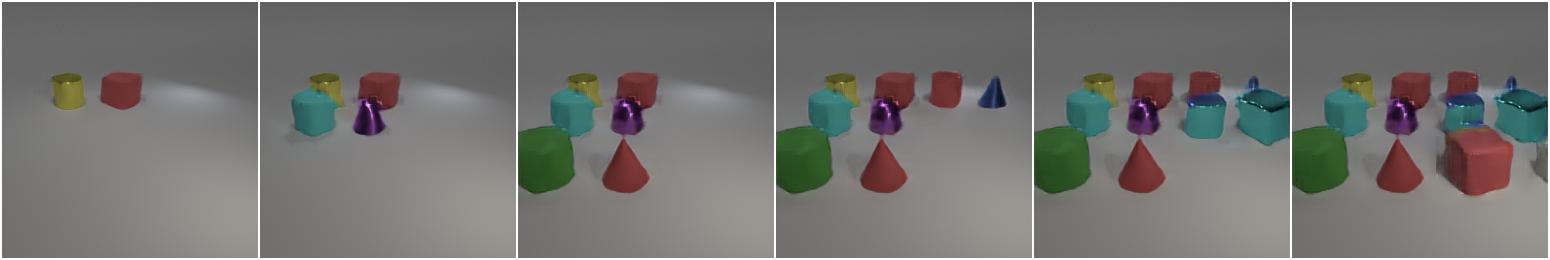}\vspace{-.1cm}
\subcaption{Add Additional Objects (Trained on Two-Object Scenes)}  
\label{subfig:add-object}
\end{subfigure}
\begin{subfigure}[t]{1.\linewidth}
  \includegraphics[width=1.\linewidth]{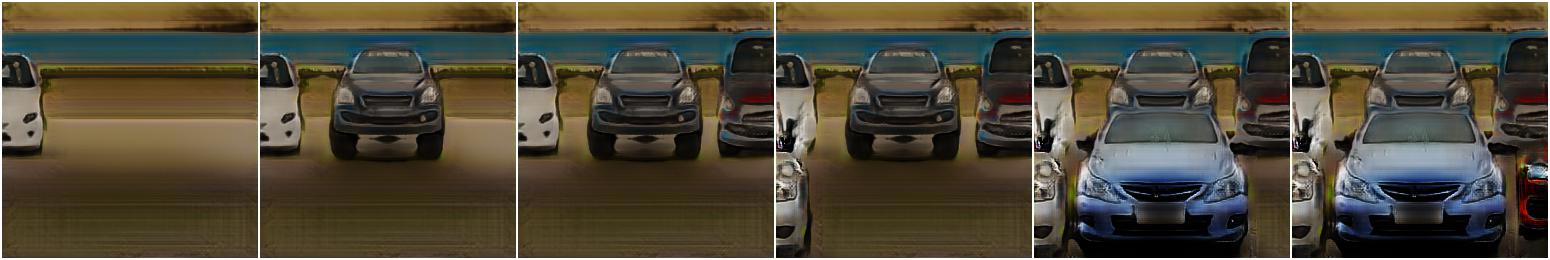}\vspace{-.1cm}
\subcaption{Add Additional Objects (Trained on Single-Object Scenes)}  
\label{subfig:add-object}
\end{subfigure}
  \vspaceunderfig\vspace{-.1cm}
  \caption{
      \textbf{Generalization Beyond Training Data.}
      As individual objects are correctly disentangled, our model allows for generating out of distribution samples at test time.
      For example, we can increase the translation ranges or add more objects than there were present in the training data.
      }
  \label{fig:generalization}
\end{figure}
\vspace{-.13cm}\boldparagraphnovspace{Disentangled Scene Generation} 
We first analyze to which degree our model learns to generate disentangled scene representations. 
In particular, we are interested if objects are disentangled from the background.
Towards this goal, we exploit the fact that our composition operator is a simple addition operation (Eq.~\ref{eq:composition}) and render individual components and object alpha maps (Eq.~\ref{eq:render-final-c}).
Note that while we always render the feature image at $16^2$ during training, we can choose arbitrary resolutions at test time.

\figref{fig:disentanglement} suggests that our method disentangles objects from the background.
Note that this disentanglement emerges without any supervision, and the model learns to generate plausible backgrounds without ever having seen a pure background image, implicitly solving an inpainting task.
We further observe that our model correctly disentangles individual objects when trained on multi-object scenes with fixed or varying number of objects.
We further find that unsupervised disentanglement is a property of our model which emerges already at the very beginning of training (\figref{fig:training-progression}).
Note how our model synthesizes individual objects before spending capacity on representing the background.

\boldparagraph{Controllable Scene Generation}
As individual components of the scene are correctly disentangled, we analyze how well they can be controlled. 
More specifically, we are interested if individual objects can be rotated and translated, but also how well shape and appearance can be controlled.
In~\figref{fig:controllable-scene-gen}, we show examples in which we control the scene during image synthesis.
We rotate individual objects, translate them in 3D space, or change the camera elevation.
By modeling shape and appearance for each entity with a different latent code, we are further able to change the objects' appearances without altering their shape.

\boldparagraph{Generalization Beyond Training Data}
The learned compositional scene representations allow us to generalize outside the training distribution.
For example, we can increase the translation ranges of objects or add more objects than there were present in the training data (\figref{fig:generalization}).

\subsection{Comparison to Baseline Methods}\vspace{-0.12cm}
\begin{table}
  \centering
  \resizebox{1.\linewidth}{!}{
  \begin{tabular}{lcccccc}
    \toprule
     & Cats & CelebA & Cars & Chairs & Churches  \\
    \midrule
    2D GAN \cite{Mescheder2018ICML}  & 18 & 15 & \textbf{16}& 59 & 19  \\
    Plat.~GAN \cite{Henzler2019ICCV}  & 318 & 321 & 299 & 199  &  242  \\
    BlockGAN \cite{Nguyen-Phuoc2020NEURIPSp} & 47 & 69 & 41& 41  & 28 \\
    HoloGAN \cite{Nguyen-Phuoc2019ICCV} & 27 & 25 & 17& 59  & 31  \\
    GRAF~\cite{Schwarz2020NEURIPS} & 26 & 25 & 39& 34  & 38  \\
    Ours  & \textbf{8} & \textbf{6} & \textbf{16}& \textbf{20} & \textbf{17} \\
    \bottomrule
\end{tabular}}
  \vspaceundertab
  \caption{
      \textbf{Quantitative Comparison.}
      We report the FID score ($\downarrow$) at $64^2$ pixels for baselines and our method.
      }
  \label{tab:tab64}
\end{table}
\begin{table}
  \centering
  \resizebox{1.\linewidth}{!}{
  \begin{tabular}{lccccc}
    \toprule
        & CelebA-HQ & FFHQ & Cars & Churches & Clevr-2 \\
    \midrule
    HoloGAN \cite{Nguyen-Phuoc2019ICCV}  & 61 & 192 & 34 & 58 & 241 \\
    \hspace{.2cm} w/o 3D~Conv  & 33 & 70 & 49 & 66 & 273 \\
    GRAF~\cite{Schwarz2020NEURIPS}  & 49 & 59 & 95 & 87 & 106 \\
    Ours & \textbf{21} & \textbf{32} & \textbf{26} & \textbf{30} & \textbf{31} \\
    \bottomrule
\end{tabular}}
  \vspaceundertab
  \caption{
    \textbf{Quantitative Comparison.}
    We report the FID score ($\downarrow$) at $256^2$ pixels for the strongest 3D-aware baselines and our method.
      }
  \label{tab:tab256}
\end{table}
\begin{figure}
  \centering
  \begin{subfigure}[t]{\linewidth}
    \centering
    \includegraphics[width=\linewidth]{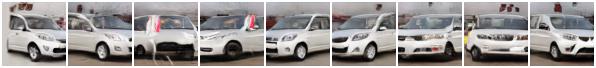}
    \includegraphics[width=\linewidth]{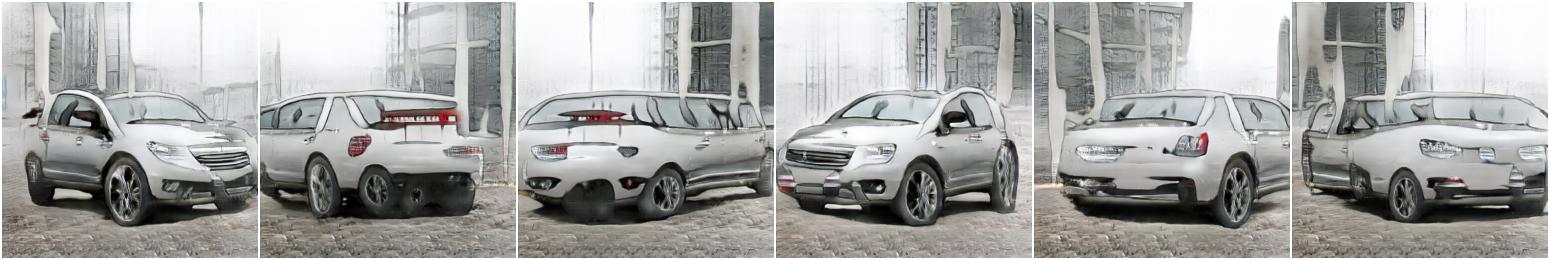}\vspace{-0.1cm}
    \caption{$360^\circ$ Object Rotation for HoloGAN~\cite{Nguyen-Phuoc2019ICCV}}
  \end{subfigure}
  \begin{subfigure}[t]{\linewidth}
    \centering
  \includegraphics[width=\linewidth]{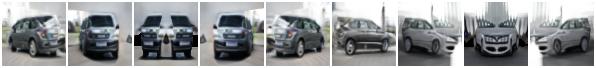}
  \includegraphics[width=\linewidth]{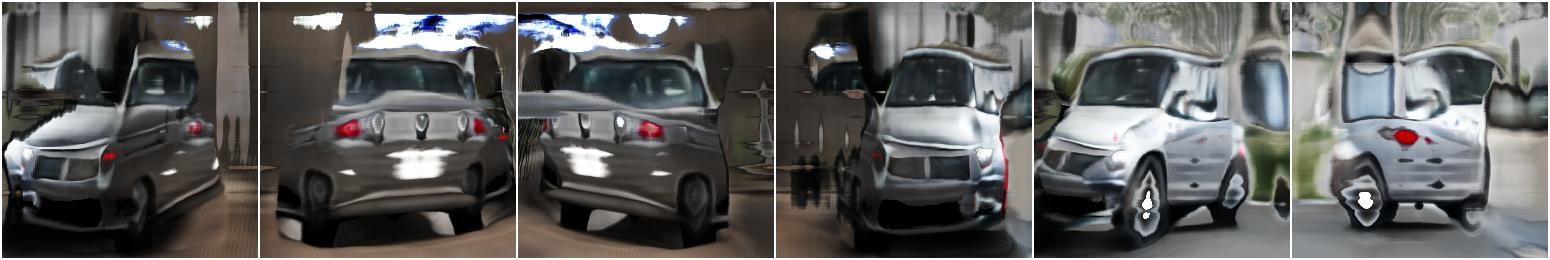}\vspace{-0.1cm}
  \caption{$360^\circ$ Object Rotation for GRAF~\cite{Schwarz2020NEURIPS}}
\end{subfigure}
  \begin{subfigure}[t]{\linewidth}
    \centering
  \includegraphics[width=\linewidth]{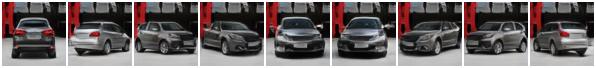}
  \includegraphics[width=\linewidth]{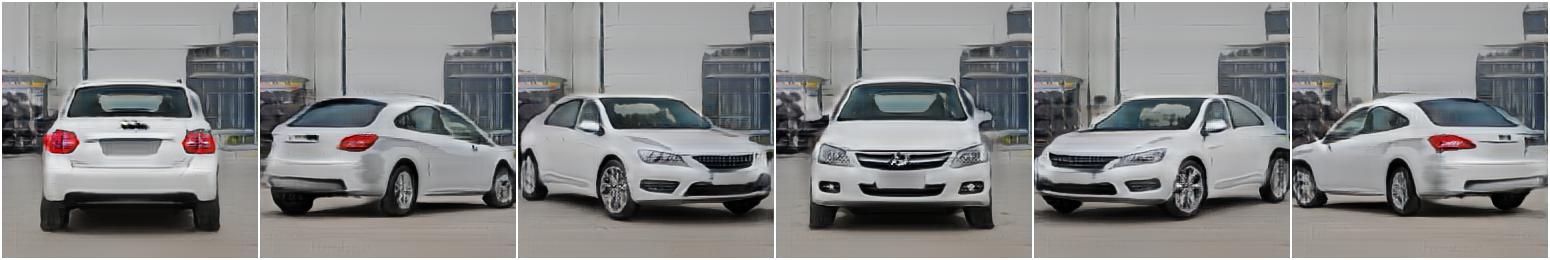}\vspace{-0.1cm}
  \caption{$360^\circ$ Object Rotation for Our Method}
\end{subfigure}
\vspaceunderfig\vspace{-.1cm}
\caption{
  \textbf{Qualitative Comparison.}
      Compared to baseline methods, we achieve more consistent image synthesis for complex scenes with cluttered background at $64^2$ (top rows) and $256^2$ (bottom rows) pixel resolutions.
      Note that we disentangle the object from the background and are able to rotate only the object while keeping the background fixed.
      }
  \label{fig:qualitative-comparison}
\end{figure}
\begin{table}
    \center
    \resizebox{1.\linewidth}{!}{
    \begin{tabular}{cccccc}
    \toprule
       2D~GAN & Plat.~GAN & BlockGAN & HoloGAN & GRAF & Ours  \\
    \midrule
    $1.69$ & $381.56$ & $4.44$ & $7.80$ & $0.68$ & $0.41$ \\
    \bottomrule
\end{tabular}}
    \vspace{-.23cm}
    \caption{
        \textbf{Network Parameter Comparison.}
        We report the number of generator network parameters in million.
    }
    \label{tab:netsize}
\end{table}
Comparing to baseline methods, our method achieves similar or better FID scores at both $64^2$~(\tabref{tab:tab64}) and $256^2$~(\tabref{tab:tab256}) pixel resolutions.
Qualitatively, we observe that while all approaches allow for controllable image synthesis on datasets of limited complexity, results are less consistent for the baseline methods on more complex scenes with cluttered backgrounds.
Further, our model disentangles the object from the background, such that we are able to control the object independent of the background (\figref{fig:qualitative-comparison}).

We further note that our model achieves similar or better FID scores than the ResNet-based 2D~GAN~\cite{Mescheder2018ICML} despite fewer network parameters ($0.41$m compared to $1.69$m).
This confirms our initial hypothesis that using a 3D representation as inductive bias results in better outputs.
Note that for fair comparison, we only report methods which are similar \wrt network size and training time (see~\tabref{tab:netsize}). %

\subsection{Ablation Studies}
\begin{table}
  \centering
  \resizebox{1.\linewidth}{!}{
  \begin{tabular}{c|cccc}
    \toprule
    Full  & -Skip & -Act.\ & +NN.\ RGB Up.\ & +Bi.\ Feat.\ Up.\ \\
    \midrule
    \textbf{16.16} & 16.66 &  21.61 & 17.28 & 20.68 \\
    \bottomrule
\end{tabular}}
  \vspaceundertab
  \caption{
      \textbf{Ablation Study.} 
      We report FID~($\downarrow$) on \textit{CompCars}
      without RGB~skip~connections (-Skip), without final activation (-Act.), with nearest neighbor instead of bilinear image upsampling (+ NN.\ RGB Up.), and with bilinear instead of nearest neighbor feature upsampling (+ Bi.\ Feat.\ Up.).
  }
  \label{tab:ablation-study}
\end{table}
\begin{figure}
  \centering
  \includegraphics[width=1.\linewidth]{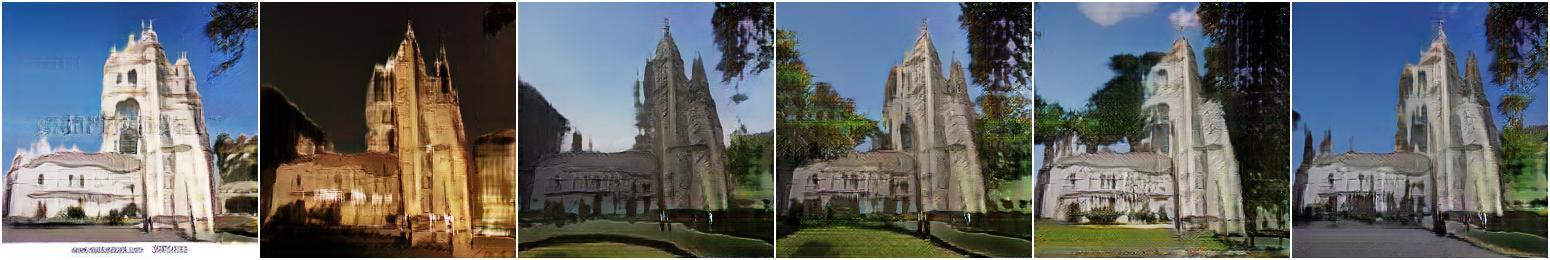}
  \includegraphics[width=1.\linewidth]{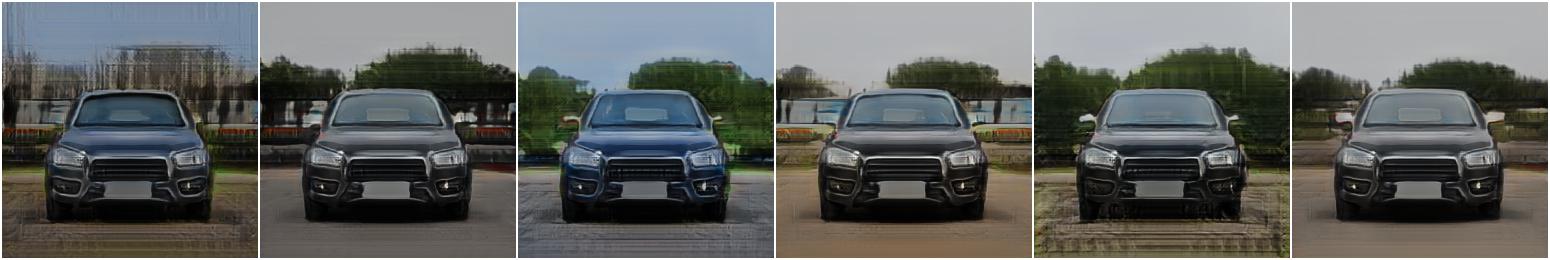}
  \vspace{-.3cm}
  \vspaceunderfig\vspace{-.1cm}
  \caption{
      \textbf{Neural Renderer.}
      We change the background while keeping the foreground object fixed for our method at $256^2$ pixel resolution.
      Note how the neural renderer realistically adapts the objects' appearances to the background.
      }
  \label{fig:neural-renderer}
\end{figure}
\begin{figure}
  \centering
  \begin{subfigure}[t]{1.\linewidth}
    \centering
  \includegraphics[width=1.\linewidth, trim= .3cm .5cm .3cm .0cm]{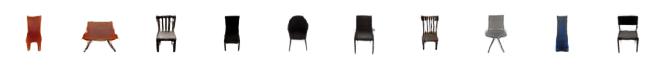}
  \subcaption{$0^\circ$ Rotation for Axis-Aligned Positional Encoding~\cite{Mildenhall2020ECCV}}
\end{subfigure}
  \begin{subfigure}[t]{1.\linewidth}
    \centering
  \includegraphics[width=1.\linewidth, trim= .3cm .5cm .3cm .0cm]{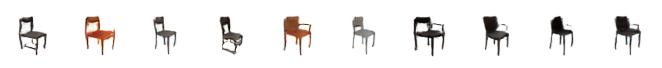}
    \subcaption{$0^\circ$ Rotation for Random Fourier Features~\cite{Tancik2020NEURIPS}}
\end{subfigure}
\vspaceunderfig\vspace{-.1cm}
  \caption{
      \textbf{Canonical Pose.}
        In contrast to random Fourier features~\cite{Tancik2020NEURIPS}, 
        axis-aligned positional encoding~\eqref{eq:pos-enc} encourages the model to learn objects in a canonical pose. %
      }
  \label{fig:canonical-pose}
\end{figure}
\vspace{-.13cm}\boldparagraphnovspace{Importance of Individual Components} The ablation study in~\tabref{tab:ablation-study} shows that our design choices of RGB skip connections, final activation function, and selected upsampling types improve results and lead to higher FID scores.

\boldparagraph{Effect of Neural Renderer}
A key difference to~\cite{Schwarz2020NEURIPS} is that we combine volume with neural rendering.
The quantitative~(\tabref{tab:tab64}~and~\ref{tab:tab256}) and qualitative comparisons~(\figref{fig:qualitative-comparison}) indicate that our approach leads to better results, in particular for complex, real-world data.
Our model is more expressive and can better handle the complexity of real scenes, \eg note how the neural renderer realistically adapts object appearances to the background (\figref{fig:neural-renderer}).
Further, we observe a rendering speed up: compared to~\cite{Schwarz2020NEURIPS}, total rendering time is reduced from $110.1$ms to $4.8$ms, and from $1595.0$ms to $5.9$ms for $64^2$ and $256^2$ pixels, respectively.

\boldparagraph{Positional Encoding} We use axis-aligned positional encoding for the input point and viewing direction (Eq.\ \ref{eq:pos-enc}). 
Surprisingly, this encourages the model to learn canoncial representations as it introduces a bias to align the object axes with highest symmetry with the canonical axes which allows the model to exploit object symmetry (\figref{fig:canonical-pose}).

\subsection{Limitations} 
\begin{figure}
  \centering
  \includegraphics[width=1.\linewidth]{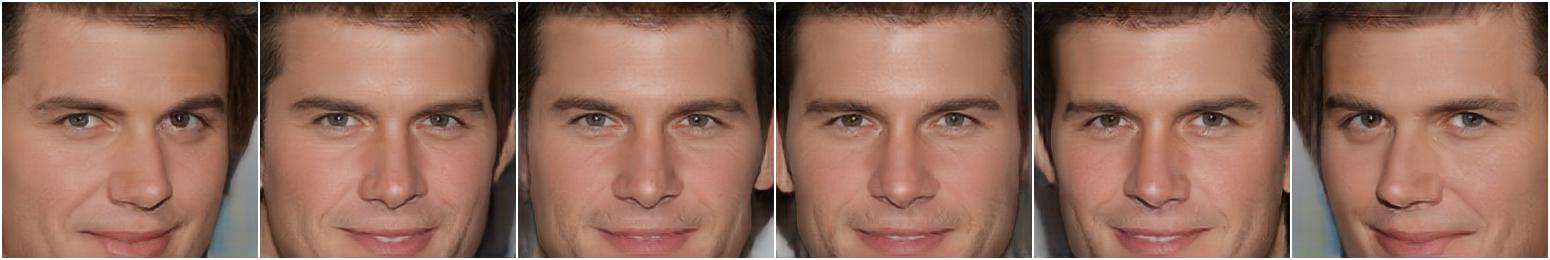}
  \includegraphics[width=1.\linewidth]{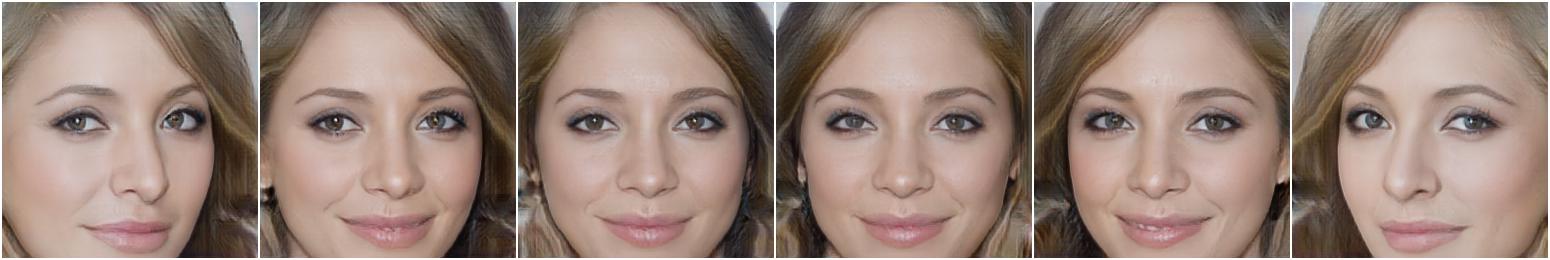}
  \vspaceunderfigext
  \caption{
      \textbf{Dataset Bias.}
        Eye and hair rotation are examples for dataset biases: 
        They primarily face the camera, and our model tends to entangle them with the object rotation. %
      }
  \label{fig:dataset-bias}
\end{figure}
\vspace{-.12cm}\boldparagraphnovspace{Dataset Bias} Our method struggles to disentangle factors of variation if there is an inherent bias in the data.
We show an example in~\figref{fig:dataset-bias}: In the celebA-HQ dataset, the eye and hair orientation is predominantly pointing towards the camera, regardless of the face rotation.
When rotating the object, the eyes and hair in our generated images do not stay fixed but are adjusted to meet the dataset bias.

\boldparagraph{Object Transformation Distributions}
We sometimes observe disentanglement failures, \eg for \textit{Churches} where the background contains a church, or for \textit{CompCars} where the foreground contains background elements (see Sup.\ Mat.).
We attribute these to mismatches between the assumed uniform distributions over camera poses and object-level transformations and their real distributions.

\section{Conclusion}\vspace{-0.15cm}

We present \textit{GIRAFFE}, a novel method for controllable image synthesis.
Our key idea is to incorporate a compositional 3D scene representation into the generative model.
By representing scenes as compositional generative neural feature fields, we disentangle individual objects from the background as well as their shape and appearance without explicit supervision.
Combining this with a neural renderer yields fast and controllable image synthesis.
In the future, we plan to investigate how the distributions over object-level transformations and camera poses can be learned from data.
Further, incorporating supervision which is easy to obtain, \eg predicted object masks, is a promising approach to scale to more complex, multi-object scenes.

\section*{Acknowledgement}\vspace{-0.17cm}
This work was supported by an NVIDIA research gift.
We thank the International Max Planck Research School for Intelligent Systems (IMPRS-IS) for supporting MN.
AG was supported by the ERC Starting Grant LEGO-3D (850533) and DFG EXC number 2064/1 - project number 390727645.

\FloatBarrier

{\small
\bibliographystyle{ieee_fullname}
\bibliography{bibliography_long,bibliography,bibliography_custom}
}

\end{document}